\documentclass{article}

\usepackage{amsmath,amssymb}
\usepackage{algorithm}
\usepackage{algpseudocode}

 \usepackage[sglblindworkshop, final]{neurips_2025}

\makeatletter
\@ifundefined{@neuripsyear}{\def\@neuripsyear{2025}}{}
\if@neuripsfinal
  \renewcommand{\@noticestring}{%
    ML For Systems workshop at Neural Information Processing Systems
    (NeurIPS \@neuripsyear).%
  }
\fi
\makeatother

\workshoptitle{MLForSys2025}

\usepackage[utf8]{inputenc} 
\usepackage[T1]{fontenc}    
\usepackage{hyperref}       
\usepackage{url}            
\usepackage{booktabs}       
\usepackage{amsfonts}       
\usepackage{nicefrac}       
\usepackage{microtype}      
\usepackage{xcolor}         
\usepackage{tabularx} 
\usepackage{multirow}
\usepackage{pifont} 
\usepackage{graphicx}

\PassOptionsToPackage{numbers,sort&compress}{natbib}
\usepackage{natbib}
\usepackage{subcaption}
\newcommand{\cmark}{\ding{51}}%
\title{Attention-Informed Surrogates for Navigating Power-Performance Trade-offs in HPC}

\author{%
Ashna Nawar Ahmed$^{1}$ \quad Banooqa Banday$^{1}$ \quad Terry Jones$^{2}$ \quad Tanzima Z. Islam$^{1}$ \\
$^{1}$Texas State University \quad $^{2}$Oak Ridge National Laboratory \\
\texttt{ashna.ahmed@txstate.edu, banooqa@txstate.edu, trjones@ornl.gov, tanzima@txstate.edu}
}

\begin{document}

\maketitle

\begin{abstract}
High-Performance Computing (HPC) schedulers must balance user performance with facility-wide resource constraints. The task boils down to selecting the optimal number of nodes for a given job. We present a surrogate-assisted multi-objective Bayesian optimization (MOBO) framework to automate this complex decision. Our core hypothesis is that surrogate models informed by attention-based embeddings of job telemetry can capture performance dynamics more effectively than standard regression techniques. We pair this with an intelligent sample acquisition strategy to ensure the approach is data-efficient. On two production HPC datasets, our embedding-informed method consistently identified higher-quality Pareto fronts of runtime-power trade-offs compared to baselines. Furthermore, our intelligent data sampling strategy drastically reduced training costs while improving the stability of the results. To our knowledge, this is the first work to successfully apply embedding-informed surrogates in a MOBO framework to the HPC scheduling problem, jointly optimizing for performance and power on production workloads.

\end{abstract}

\section{Introduction}
Exascale HPC systems face a fundamental trade-off: minimizing job runtime to accelerate science while reducing power consumption for sustainability. This tension plays out in everyday decisions--scientists must choose how many nodes to request, often guessing between too few (risking termination) and too many (wasting energy and resources). Traditional scheduling heuristics, such as First-Come, First-Served with backfilling, optimize for a single objective (most commonly runtime) and lack the flexibility to adapt to workload variability or guide resource selection.

Recent work in HPC scheduling has focused on runtime prediction using ML models \citep{ramachandran2024asoc,islam2016a,dey2025modelx,islam2019toward,ramadan2021comparative,ramadan2023novel,yeom2016data,dey2024relative,nichols2024predicting}, single-objective reinforcement learning \citep{fan2021deepreinforcementagentscheduling, fan2021dras, wang2021rlschert, zhang2019rlscheduler, souza2024hpc_rl}, or offline parameter tuning using surrogate-based Bayesian optimization (BO) \citep{bhatele2020hiperbot, eriksson2021scalable3bo, kim2022bofss, dorier2022storage_autotune, perez2025parallel_autotune}. While these approaches show promise, they optimize a single objective, assume full access to clean input data, and do not expose runtime–power trade-offs. Because runtime and power are inherently conflicting objectives, it is important to capture the trade-off space, as no one solution may optimize both. As such, power-performance trade-off modeling can be formulated as a multi-objective optimization problem. 
The MOBO literature and acquisition strategies \citep{daulton2020qehvi}, improve sample efficiency. However, these methods have not been explored in the context of real-world and noisy HPC telemetry data, characterized by missing entries, inconsistent counters, and large variability across jobs. This observation reveals a key limitation in prior work: the lack of methods for modeling conflicting objectives that operate effectively on large-scale, noisy, and irregular telemetry data.

HPC telemetry presents three core challenges: (1) it is massive, noisy, and irregular, making surrogate training unstable \citep{hpclog}; we address this challenge with \textbf{intelligent sample acquisition} to guide selective sampling; (2) it is multimodal and variable in scale, with log-transformed targets amplifying sensitivity to noise \citep{Arik_Pfister_2021}; in response, we propose \textbf{attention-based embeddings} to highlight informative features and suppress redundancy; and (3) runtime and power are conflicting objectives, which traditional heuristics cannot jointly optimize; we model the power-performance trade-off by leveraging the \textbf{MOBO} method. In practice, our use of MOBO is intended as a decision-support tool: it can recommend the number of nodes a user should request for a job, a choice that is otherwise difficult to make manually and often leads to suboptimal runtime or wasted power.

These challenges motivate two hypotheses tested in this study:
\textbf{H1:} attention-based embeddings improve surrogate quality compared to direct regressors.
\textbf{H2:} MOBO captures runtime–power trade-offs more effectively than single-objective BO (SOBO) or random baselines.

\textbf{Contributions:} In this work, we (1) Propose a surrogate-driven MOBO framework that integrates attention-based embeddings and intelligent sample acquisition for robust modeling from irregular HPC telemetry;
(2) Demonstrate across two large-scale HPC datasets that both \textbf{H1} and \textbf{H2} hold consistently;
(3) Provide a reproducible pipeline for surrogate training and optimization that explicitly balances 
runtime and power trade-offs 
for decision support.


\section{Related Work}
There exists a large body of work that leverages ML for predicting performance for HPC systems using single metric performance prediction~\cite{islam2016a,dey2025modelx,islam2019toward,ramadan2021comparative,ramadan2023novel,yeom2016data,dey2024relative,nichols2024predicting}. More recently, Ramachandran et al. \citep{ramachandran2024asoc} propose a hybrid Genetic Algorithm (GA) and ML approach for runtime prediction. Their method replaces inaccurate user-supplied runtimes with GA-defined runtime classes, enabling models such as KNN, SVR, XGBoost, and DNN to achieve R² > 0.8. 
While effective for runtime prediction, this work does not incorporate power modeling or multi-objective trade-offs.

In parallel, the optimization community has advanced the MOBO methodology to handle conflicting objectives by learning surrogate models and guiding the search toward Pareto-optimal trade-offs. Daulton et al.\citep{daulton2020qehvi} introduced qEHVI, a differentiable acquisition function that efficiently improves hypervolume in parallel MOBO settings. Such methods aim to approximate the \emph{Pareto front}, the set of solutions where no objective (e.g., runtime) can be improved without worsening another (e.g., power). Pareto optimization is valuable because it exposes trade-offs directly, giving decision-makers multiple balanced options rather than a single outcome\citep{deb}. However, these approaches have not yet been applied to HPC job scheduling.



Jannach et al.~\citep{jannach2023multiobjectiveRecsys} survey multi-objective recommender systems and highlight the role of Pareto optimization~\citep{deb} in balancing competing goals such as accuracy and diversity. Similar to their ranking of items for end users, our framework produces ranked recommendations of candidate resource configurations in HPC scheduling. The difference lies in the domain: they target user-facing recommendation systems, while we focus on runtime–power trade-offs in supercomputing workloads. 

Surrogate-based Bayesian optimization (BO) has been widely applied to offline autotuning in HPC, including compiler and thread parameters~\citep{bhatele2020hiperbot}, scalable asynchronous searches~\citep{eriksson2021scalable3bo}, file system and storage tuning~\citep{kim2022bofss, dorier2022storage_autotune}, and large-scale parallel optimization~\citep{perez2025parallel_autotune}. These approaches improve subsystem efficiency but remain single-objective and do not capture conflicting runtime–power trade-offs.
Our work fills these gaps by using attention to improve surrogate model quality on irregular HPC telemetry, and applying these models in MOBO to find better runtime–power trade-offs. 

\section{Our Approach}
We design a surrogate-driven MOBO framework to address the scale, heterogeneity, and conflicting objectives of HPC scheduling.  
First, to cope with massive but irregular telemetry, we adopt intelligent sample acquisition based on active learning (see Appendix: Intelligent Sample Acquisition Overview). Instead of training on all raw logs, the surrogate iteratively selects the most informative samples, improving efficiency and robustness in the presence of noisy data.  

To handle heterogeneity in HPC telemetry, we use attention-based embeddings to highlight informative features and suppress irrelevant or redundant inputs. We implement this using TabNet \citep{huang2020tabtransformer}, which learns feature-level attention directly from structured data. We hypothesize that such embeddings enable lightweight regressors--such as Random Forest \citep{breiman2001randomforest}, XGBoost \citep{chen2016xgboost}, and LightGBM \citep{ke2017lightgbm}--to generalize better, train faster, and remain more interpretable. While transformer-based models have shown promise for modality fusion and long-range dependencies \citep{vaswani2017attention,dosovitskiy2021vit,jaegle2021perceiver}, they are resource-intensive, require large training sets, and lack interpretability, which limits their practicality for deployment in HPC systems. 

Finally, to model runtime-power trade-off space, we integrate trained surrogates into the MOBO framework. This allows us to capture runtime–power trade-offs explicitly, predict performance for candidate node allocations, and generate Pareto fronts that quantify the balance between runtime and power consumption. In HPC, this often involves balancing runtime and power consumption--improving one typically worsens the other. The Pareto front captures all such non-dominated solutions, providing a spectrum of efficient choices for informed decision-making.
By comparing against single-objective and random baselines, we show that the MOBO framework produces higher-quality Pareto fronts--capturing more balanced runtime–power trade-offs.

End-to-end, as illustrated in Fig.~\ref{fig:pipeline} in the Supplementary Material section, the pipeline consists of pre-processing HPC datasets, selecting informative samples, extracting embeddings, fitting surrogate models, integrating them into the MOBO framework, predicting runtime and power outcomes, and visualizing Pareto fronts for decision-making in HPC scheduling.

\section{Experimental Setup}
\textbf{Datasets.} We evaluate on two real-world HPC job-log datasets: PM100~\citep{pm100}, which contains 231{,}238 job records with 35 features, and Adastra~\citep{adastra-cines-2024}, which contains 15{,}285 job records with 35 features. Both datasets include categorical fields (e.g., queue, partition), numerical metrics (e.g., runtime, memory, power), and system-level telemetry derived from production supercomputers.
\textbf{Pre-processing.}
We pre-process these datasets including handling missing values, aggregating node-level power to job-level totals, and convert categorical, numeric, and time-series–derived signals (e.g., GPU bursts, I/O spikes) into a single structured input for modeling.
\textbf{Surrogate Models.} We compare two classes of surrogates: (1) Transformer-based models using TabNet, and (2) light-weight regressors--Random Forest \citep{breiman2001randomforest}, XGBoost \citep{chen2016xgboost}, and LightGBM \citep{ke2017lightgbm} trained using attention-based embeddings. All surrogates are embedded in a MOBO loop that generates Pareto fronts capturing runtime–power trade-offs.
\textbf{Baselines.} We compare against three baselines: (1) SOBO (runtime-only), (2) SOBO (power-only), and (3) Random Search. SOBO quantifies the impact of optimizing one objective in isolation, while Random estimates the value of surrogate-guided acquisition.
\textbf{Compute Environment.} Experiments are run on the Stampede3 supercomputer at TACC \citep{tacc-stampede3} using CPU nodes for training and MOBO evaluations. Hyperparameter and training details are in Tables~\ref{tab:common_settings} and~\ref{tab:pipeline_hparams}. 

\section{Results}

\begin{minipage}{\linewidth}
\centering
\small
\captionof{table}{Summary of results supporting Hypotheses 1 and 2, and dataset sizes under sampling.}
\label{tab:results_new}
\begin{tabularx}{\linewidth}{l| l X X p{2.3cm}}
\toprule
\textbf{Dataset} & \textbf{Metric} & \textbf{Hypothesis 1 (Embeddings vs. TabNet)} & \textbf{Hypothesis 2 (MOBO vs. Baselines)} & \textbf{Sample size} \\
\midrule
\multirow{2}{*}{PM100}
  & HV     & \cmark\ Embeddings have \emph{orders of magnitude} higher HV than Regressor
           & \cmark\ Improved HV 24\% vs.\ SOBO--Runtime (TabNet Regressor) 
           & 50\%$\rightarrow$73,983; 75\%$\rightarrow$77,278; 100\%$\rightarrow$109,202 \\ \cline{2-5}
  & Spread & \cmark\ Embeddings have $\sim$99\% lower Spread than Regressor
           & \cmark\ MOBO best in 3/4 families (75\%) 
           & Original$\rightarrow$231,238 \\
\midrule
\multirow{2}{*}{Adastra}
  & HV     & \cmark\ Embeddings have 37\% more HV than Regressor
           & \cmark\ Improved HV 37\% vs.\ SOBO--Runtime (TabNet Regressor) 
           & 50\%$\rightarrow$3,964; 75\%$\rightarrow$4,163; 100\%$\rightarrow$4,547 \\\cline{2-5}
  & Spread & \cmark\ Embeddings have $\sim$90\% lower Spread than Regressor
           & \cmark\ MOBO best in 3/4 families (75\%) 
           & Original$\rightarrow$15,285 \\
\bottomrule
\end{tabularx}
\end{minipage}


\textbf{Impact of Attention-Based Embeddings.}
This experiment tests \textbf{H1} by evaluating whether attention-based embeddings improve surrogate quality by enabling simpler models to generalize better. 
\underline{Observations.}
Table~\ref{tab:results_new} and Fig.~\ref{fig:h1} (Supplementary Material section) show that embedding-informed simple surrogate models outperform a transformer architecture across both datasets in most settings, supporting \textit{H1}. Simpler, tree-based models outperform transformer-based models in our setting because subsampling significantly reduces dataset size, limiting the effectiveness of deep architectures that require large amounts of data to generalize well. Moreover, we study the impact of these high quality surrogates on the downstream pipeline and find that: (1) On PM100, embeddings reduce Spread by $\sim$99\% and yield orders-of-magnitude higher HV under SOBO; (2) On Adastra, they improve HV by 37\% and reduce Spread by $\sim$90\%. These results can be explained by the fact that attention mechanism isolates relevant features while filtering out noise, making 
tree-based models to train faster and remain robust. 

\textbf{Impact of Intelligent Sample Acquisition.}
This experiment evaluates whether intelligent sample acquisition improves surrogate stability and optimization quality for HPC job scheduling. 
%
\underline{Observations.}
As summarized in Table~\ref{tab:results_new}, embedding-informed surrogates consistently outperform TabNet regressors in both HV and spread, supporting \textit{H1}. MOBO generally outperforms SOBO and Random, with clearer gains under sampling (H2). 
%
Across both PM100 and Adastra, intelligent sample acquisition enables stable and accurate surrogates using only 50–75\% of the data, significantly reducing training overhead. At 50\%, both datasets maintain consistent hypervolume (HV) and spread; at 75\%, surrogate accuracy (MAPE $\approx$ 0.99) plateaus (Supplementary Material section, Table~\ref{tab:effect}).
Compared to full-data training, intelligent sample acquisition yields faster convergence and lower variance (Supplementary Material section: Fig.~\ref{fig:active2}, \ref{fig:active}). On PM100, it improves Pareto diversity: spread drops from $2.5\times10^5$ to below $10^4$. On Adastra, it suppresses erratic HV values (from $10^{16}$ to $10^{13}$), producing smoother trade-offs.

\textbf{Impact on Timing Overheads.}
This experiment evaluates whether intelligent sample acquisition reduces the computational cost of surrogate-driven MOBO by lowering runtime overhead across preprocessing, training, and optimization stages.
\underline{Observations.}
Table~\ref{tab:timing_overhead} (Supplementary Material section) shows that sampling consistently reduces total execution time. On PM100, runtime drops from 6480s to 6009s, with the largest savings in surrogate training. On Adastra, the reduction is more pronounced—from 2699s to 1962s (27\% decrease). While MOBO evaluation time increases slightly due to repeated acquisition steps, this cost is offset by the reduction in data volume. The dominant gains come from fewer training samples, which shorten preprocessing and model training.
These results reinforce that intelligent sampling not only improves surrogate stability and optimization quality, but also reduces end-to-end runtime. 

\textbf{Limitations and Future Work.}
Our present study evaluates two production traces and focuses on runtime-power trade-offs; going forward we will (i) \emph{deployment \& latency}: profile end-to-end overheads by reporting surrogate inference time and BO cycle wall-time, and tighten the loop via cached embeddings and warm-started retraining to meet scheduler budgets; (ii) \emph{scalability in objectives}: extend beyond two objectives and assess complexity/quality trade-offs using scalarization warm-starts and NEHVI/log-NEHVI variants with controlled candidate sets; (iii) \emph{model choice}: examine when lightweight regressors suffice versus transformer-based surrogates under different data regimes, with an emphasis on stability, small-sample behavior, and compute/latency footprints; and (iv) \emph{generalization across systems}: study portability across clusters via re-embedding or adapter heads with minimal calibration runs, and quantify the impact of observed domain gaps. These steps concentrate on practical readiness and breadth while preserving the simplicity that makes the current pipeline deployable.


\textbf{Broader Impacts.}
This research will have a significant impact on computational scientists by automating the complex decision of how many nodes to use for their HPC jobs. By embedding this intelligence into the scheduler, users are freed from this error-prone task, allowing them to focus on science. Building on our previous work~\citep{dey2025modelx}, which shows that accurate runtime prediction can cut time-to-science by 71\% and resource usage by 42\%, we can extend the work further to balance runtime with power consumption. Such a capability will accelerate scientific discovery, boost the efficiency of multi-million-dollar HPC systems, and lower their operational cost.

\section{Conclusion}


This work
introduces an end-to-end framework for data-efficient, multi-objective decision-making in HPC scheduling. We improve the deployability of AI-driven performance models by using intelligent sample acquisition for data reduction and attention-based embeddings to improve prediction accuracy. These enhanced surrogates serve as the foundation for a MOBO-based optimization engine that effectively models the power-performance trade-off space. The result is a scalable performance modeling methodology that can be integrated into next-generation HPC schedulers to automate complex configuration decisions and optimize multiple operational goals simultaneously.

\section{Acknowledgement}
This material is based upon work supported by the U.S. Department of Energy, Office of Science under Award Number DE-SC0022843. Additional support was provided by the U.S. Department of Energy, Office of Science, Office of Advanced Scientific Computing Research, Next-Generation Scientific Software Technologies program under Contract Number DE-AC05-00OR22725.


\begingroup
\small
\bibliographystyle{unsrt} 
\bibliography{bibliography}

@misc{fan2021deepreinforcementagentscheduling,
      title={Deep Reinforcement Agent for Scheduling in HPC}, 
      author={Yuping Fan and Zhiling Lan and Taylor Childers and Paul Rich and William Allcock and Michael E. Papka},
      year={2021},
      eprint={2102.06243},
      archivePrefix={arXiv},
      primaryClass={cs.DC},
      url={https://arxiv.org/abs/2102.06243}, 
}

@article{fan2021dras,
  title        = {DRAS-CQSim: A Reinforcement Learning based Framework for HPC Cluster Scheduling},
  author       = {Fan, Yuping and Lan, Zhiling},
  journal      = {arXiv preprint arXiv:2105.07526},
  year         = {2021},
  url          = {https://arxiv.org/abs/2105.07526}
}

@article{wang2021rlschert,
  author    = {Wang, Qiqi and Zhang, Hongjie and Qu, Cheng and Shen, Yu and Liu, Xiaohui and Li, Jing},
  title     = {RLSchert: An HPC Job Scheduler Using Deep Reinforcement Learning and Remaining Time Prediction},
  journal   = {Applied Sciences},
  volume    = {11},
  number    = {20},
  pages     = {9448},
  year      = {2021},
  doi       = {10.3390/app11209448}
}

@inproceedings{zhang2019rlscheduler,
  author    = {Zhang, Di and Dai, Dong and He, Youbiao and Bao, Forrest Sheng and Xie, Bing},
  title     = {RLScheduler: An Automated HPC Batch Job Scheduler Using Reinforcement Learning},
  booktitle = {Proceedings of [Conference Name]},
  year      = {2019},
  pages     = {xx--xx}
}

@article{souza2024hpc_rl,
  author    = {Souza, Abel and Pelckmans, Kristiaan and Tordsson, Johan},
  title     = {A HPC Co-Scheduler with Reinforcement Learning},
  journal   = {arXiv preprint arXiv:2401.09706},
  year      = {2024},
  url       = {https://arxiv.org/abs/2401.09706}
}

@inbook{deb,
  author    = {Deb, Kalyanmoy},
  title     = {Multi-Objective Optimization Using Evolutionary Algorithms},
  publisher = {Wiley},
  year      = {2001},
  address   = {New York, NY, USA}
}

@INPROCEEDINGS{hpclog,
  author={Park, Byung H. and Hukerikar, Saurabh and Adamson, Ryan and Engelmann, Christian},
  booktitle={2017 IEEE International Conference on Cluster Computing (CLUSTER)}, 
  title={Big Data Meets HPC Log Analytics: Scalable Approach to Understanding Systems at Extreme Scale}, 
  year={2017},
  volume={},
  number={},
  pages={758-765},
  doi={10.1109/CLUSTER.2017.113}}

@article{Arik_Pfister_2021, 
title={TabNet: Attentive Interpretable Tabular Learning}, 
volume={35}, 
url={https://ojs.aaai.org/index.php/AAAI/article/view/16826}, 
DOI={10.1609/aaai.v35i8.16826}, 
number={8}, 
journal={Proceedings of the AAAI Conference on Artificial Intelligence}, 
author={Arik, Sercan Ö. and Pfister, Tomas}, 
year={2021}, 
month={May}, 
pages={6679-6687} }

@inproceedings{pm100,
author = {Antici, Francesco and Seyedkazemi Ardebili, Mohsen and Bartolini, Andrea and Kiziltan, Zeynep},
title = {PM100: A Job Power Consumption Dataset of a Large-scale Production HPC System},
year = {2023},
isbn = {9798400707858},
publisher = {Association for Computing Machinery},
address = {New York, NY, USA},
url = {https://doi.org/10.1145/3624062.3624263},
doi = {10.1145/3624062.3624263},
booktitle = {Proceedings of the SC '23 Workshops of the International Conference on High Performance Computing, Network, Storage, and Analysis},
pages = {1812–1819},
numpages = {8},
location = {Denver, CO, USA},
series = {SC-W '23}
}

@misc{adastra-cines-2024,
  title        = {{Adastra}: Log Data from the HPE-Cray EX Supercomputer at CINES},
  howpublished = {\url{https://www.top500.org/system/180047/}}, 
  year         = {2024},
  note         = {Accessed: 2025-08-18}
}

@misc{tacc-stampede3,
  author       = {{Texas Advanced Computing Center}},
  title        = {Stampede3 User Guide},
  howpublished = {\url{https://docs.tacc.utexas.edu/hpc/stampede3/}},
  year         = {2025},
  note         = {Accessed: 2025-08-18}
}

@article{ramachandran2024asoc,
  title   = {Combining Machine Learning Techniques and Genetic Algorithm for Predicting Run Times of High Performance Computing Jobs},
  author  = {Ramachandran, S. and Jayalal, M. L. and Vasudevan, M. and Das, S. and Jehadeesan, R.},
  journal = {Applied Soft Computing},
  volume  = {165},
  pages   = {112053},
  year    = {2024},
  publisher = {Elsevier},
  doi     = {10.1016/j.asoc.2024.112053}
}

@inproceedings{daulton2020qehvi,
  author    = {Daulton, Samuel and Balandat, Maximilian and Bakshy, Eytan},
  title     = {Differentiable Expected Hypervolume Improvement for Parallel Multi-Objective Bayesian Optimization},
  booktitle = {Advances in Neural Information Processing Systems 33 (NeurIPS 2020)},
  year      = {2020},
  pages     = {9851--9864},
  publisher = {Curran Associates, Inc.},
  url       = {https://proceedings.neurips.cc/paper/2020/hash/6e55d70a3eac2de52c5f0d3b3c3a9b1a-Abstract.html}
}

@article{jannach2023multiobjectiveRecsys,
  author      = {Jannach, Dietmar and Abdollahpouri, Himan},
  title       = {A Survey on Multi-Objective Recommender Systems},
  journal     = {Frontiers in Big Data},
  volume      = {6},
  pages       = {1157899},
  year        = {2023},
  doi         = {10.3389/fdata.2023.1157899}
}

@inproceedings{bhatele2020hiperbot,
  author    = {Menon, Harshitha and Bhatele, Abhinav and Gamblin, Todd},
  title     = {Auto-tuning Parameter Choices in HPC Applications using Bayesian Optimization},
  booktitle = {Proceedings of the 2020 IEEE International Parallel and Distributed Processing Symposium (IPDPS)},
  pages     = {831--840},
  year      = {2020},
  publisher = {IEEE},
  doi       = {10.1109/IPDPS47924.2020.00049}
}

@article{eriksson2021scalable3bo,
  author = {Tran, Anh},
  title  = {Scalable\textsuperscript{3}-BO: Big Data meets HPC - A scalable asynchronous parallel high-dimensional Bayesian optimization framework on supercomputers},
  journal = {arXiv preprint},
  volume  = {arXiv:2108.05969},
  year    = {2021},
  url     = {https://arxiv.org/abs/2108.05969}
}

@article{kim2022bofss,
  author    = {Kim, Kyurae and Kim, Youngjae and Park, Sungyong},
  title     = {A Probabilistic Machine Learning Approach to Scheduling Parallel Loops with Bayesian Optimization},
  journal   = {IEEE Transactions on Parallel and Distributed Systems},
  year      = {2022},
  note      = {Preprint at arXiv},
  url       = {https://arxiv.org/abs/2206.05787}
}

@article{dorier2022storage_autotune,
  author    = {Dorier, Matthieu and Egele, Romain and Balaprakash, Prasanna and Koo, Jaehoon and Madireddy, Sandeep and Ramesh, Srinivasan and Malony, Allen D. and Ross, Rob},
  title     = {HPC Storage Service Autotuning Using Variational-Autoencoder-Guided Asynchronous Bayesian Optimization},
  journal   = {arXiv preprint arXiv:2210.00798},
  year      = {2022},
  url       = {https://arxiv.org/abs/2210.00798}
}

@article{perez2025parallel_autotune,
  author    = {Perez, Adrian and Ockerman, Seth and Aikman, Tristan and Ibrahim, Khaled Z.},
  title     = {Parallelizing Autotuning for HPC Applications: Unveiling the Potential of Speculation Strategy in Bayesian Optimization},
  journal   = {The International Journal of High Performance Computing Applications},
  year      = {2025},
  note      = {In press},
}

@article{breiman2001randomforest,
  author    = {Breiman, Leo},
  title     = {Random Forests},
  journal   = {Machine Learning},
  volume    = {45},
  number    = {1},
  pages     = {5--32},
  year      = {2001},
  publisher = {Springer},
  doi       = {10.1023/A:1010933404324}
}

@inproceedings{chen2016xgboost,
  author    = {Chen, Tianqi and Guestrin, Carlos},
  title     = {XGBoost: A Scalable Tree Boosting System},
  booktitle = {Proceedings of the 22nd ACM SIGKDD International Conference on Knowledge Discovery and Data Mining (KDD '16)},
  year      = {2016},
  pages     = {785--794},
  publisher = {ACM},
  doi       = {10.1145/2939672.2939785},
  address   = {New York, NY, USA}
}

@inproceedings{ke2017lightgbm,
  author    = {Ke, Guolin and Meng, Qi and Finley, Thomas and Wang, Taifeng and Chen, Wei and Ma, Weidong and Ye, Qiwei and Liu, Tie-Yan},
  title     = {LightGBM: A Highly Efficient Gradient Boosting Decision Tree},
  booktitle = {Advances in Neural Information Processing Systems 30 (NeurIPS 2017)},
  pages     = {3146--3154},
  year      = {2017},
  url       = {https://papers.nips.cc/paper/2017/hash/6449f44a102fde848669bdd9eb6b76fa-Abstract.html}
}

@inproceedings{vaswani2017attention,
  title={Attention Is All You Need},
  author={Vaswani, Ashish and Shazeer, Noam and Parmar, Niki and Uszkoreit, Jakob and Jones, Llion and Gomez, Aidan N. and Kaiser, {\L}ukasz and Polosukhin, Illia},
  booktitle={Advances in Neural Information Processing Systems 30 (NeurIPS 2017)},
  pages={5998--6008},
  year={2017}
}

@inproceedings{dosovitskiy2021vit,
  title={An Image is Worth 16x16 Words: Transformers for Image Recognition at Scale},
  author={Dosovitskiy, Alexey and Beyer, Lucas and Kolesnikov, Alexander and Weissenborn, Dirk and Zhai, Xiaohua and Unterthiner, Thomas and Dehghani, Mostafa and Minderer, Matthias and Heigold, Georg and Gelly, Sylvain and Uszkoreit, Jakob and Houlsby, Neil},
  booktitle={International Conference on Learning Representations (ICLR)},
  year={2021}
}

@inproceedings{jaegle2021perceiver,
  title={Perceiver: General Perception with Iterative Attention},
  author={Jaegle, Andrew and Gimeno, Felix and Brock, Andrew and Vinyals, Oriol and Zisserman, Andrew and Carreira, Jo{\~a}o},
  booktitle={International Conference on Machine Learning (ICML)},
  pages={4651--4664},
  year={2021}
}

@inproceedings{huang2020tabtransformer,
  title={TabTransformer: Tabular Data Modeling Using Contextual Embeddings},
  author={Huang, Xin and Khetan, Ashish and Cvitkovic, Milan and Karnin, Zohar},
  booktitle={Proceedings of the 29th ACM International Conference on Information \& Knowledge Management},
  pages={1027--1035},
  year={2020},
  organization={ACM},
  doi={10.1145/3340531.3412002}
}

@INPROCEEDINGS{dey2025modelx,
  author={Dey, Arunavo and Antony, Neil and Dhakal, Aakash R. and Thopalli, Kowshik and Thiagarajan, Jayaraman J. and Patki, Tapasya and Marathe, Aniruddha and Scogland, Tom and Yeom, Jae-Seung and Islam, Tanzima},
  booktitle={The 34th ACM International Symposium on High-Performance Parallel and Distributed Computing (HPDC)},
  title={{ModelX: A Novel Transfer Learning Approach Across Heterogeneous Datasets}},
  year={2025},
note={Accepted (19\% acceptance rate)}
}

@inproceedings{islam2019toward,
  title={Toward a Programmable Analysis and Visualization Framework for Interactive Performance Analytics},
  author={Islam, Tanzima and Ayala, Alexis and Jensen, Quentin and Ibrahim, Khaled},
  booktitle={2019 IEEE/ACM International Workshop on Programming and Performance Visualization Tools (ProTools)},
  pages={70--77},
  year={2019},
  organization={IEEE}
}

@INPROCEEDINGS{ramadan2023novel,
  author={Ramadan, Tarek and Lahiry, Ankur and Islam, Tanzima Z.},
  booktitle={2023 International Conference on Machine Learning and Applications (ICMLA)}, 
  title={Novel Representation Learning Technique Using Graphs for Performance Analytics}, 
  year={2023},
  volume={},
  number={},
  pages={1311-1318},
  doi={10.1109/ICMLA58977.2023.00198}}

@inproceedings{ramadan2021comparative,
  title={Comparative Code Structure Analysis using Deep Learning for Performance Prediction},
  author={Ramadan, Tarek and Islam, Tanzima Z and Phelps, Chase and Pinnow, Nathan and Thiagarajan, Jayaraman J},
  booktitle={2021 IEEE International Symposium on Performance Analysis of Systems and Software (ISPASS)},
  pages={151--161},
  year={2021},
  organization={IEEE}
}

@inproceedings{islam2016a,
  title={A machine learning framework for performance coverage analysis of proxy applications},
  author={Islam, Tanzima Z and Thiagarajan, Jayaraman J and Bhatele, Abhinav and Schulz, Martin and Gamblin, Todd},
  booktitle={SC'16: Proceedings of the International Conference for High Performance Computing, Networking, Storage and Analysis},
  pages={538--549},
  year={2016},
  organization={IEEE}
}

@inproceedings{yeom2016data,
  title={Data-driven performance modeling of linear solvers for sparse matrices},
  author={Yeom, Jae-Seung and Thiagarajan, Jayaraman J and Bhatele, Abhinav and Bronevetsky, Greg and Kolev, Tzanio},
  booktitle={Performance Modeling, Benchmarking and Simulation of High Performance Computer Systems (PMBS), International Workshop on},
  pages={32--42},
  year={2016},
  organization={IEEE}
}

@INPROCEEDINGS{dey2024relative,
  author={Dey, Arunavo and Dhakal, Aakash and Islam, Tanzima Z. and Yeom, Jae-Seung and Patki, Tapasya and Nichols, Daniel and Movsesyan, Alexander and Bhatele, Abhinav},
  booktitle={2024 IEEE 48th Annual Computers, Software, and Applications Conference (COMPSAC)}, 
  title={Relative Performance Prediction Using Few-Shot Learning}, 
  year={2024},
  volume={},
  number={},
  pages={1764-1769},
  doi={10.1109/COMPSAC61105.2024.00278}}

@INPROCEEDINGS{nichols2024predicting,
  author={Nichols, Daniel and Movsesyan, Alexander and Yeom, Jae-Seung and Sarkar, Abhik and Milroy, Daniel and Patki, Tapasya and Bhatele, Abhinav},
  booktitle={2024 IEEE International Parallel and Distributed Processing Symposium (IPDPS)}, 
  title={Predicting Cross-Architecture Performance of Parallel Programs}, 
  year={2024},
  volume={},
  number={},
  pages={570-581},
  doi={10.1109/IPDPS57955.2024.00057}}
\endgroup
\clearpage





\appendix

\section{Supplementary Material}
\begin{figure}[t]
  \centering
  \includegraphics[width=\linewidth]{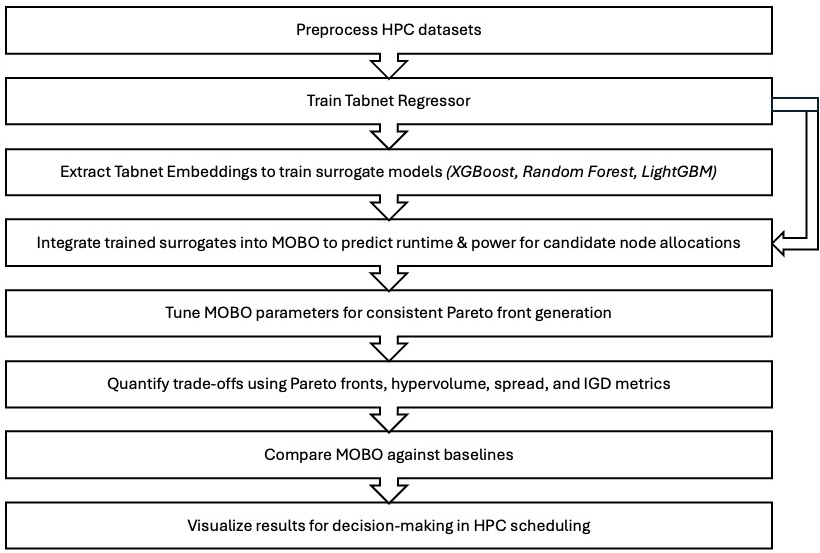}
  \caption{End-to-end pipeline for surrogate-driven MOBO on HPC scheduling.}
  \label{fig:pipeline}
\end{figure}

The full pipeline is shown in Figure \ref{fig:pipeline}.

\paragraph{Intelligent Sample Acquisition Overview}
We propose a loss-proportional subset sampler for mixed regression--classification targets.
Given a table $D=\{(x_i,y_i)\}_{i=1}^n$ partitioned into regression targets $T_r$ and classification targets $T_c$, we first construct a numeric sampling view $V$ by (i) reducing structured power arrays to scalar totals, (ii) converting datetimes to epoch seconds and deriving durations (wait, queue, job duration), (iii) casting string numerics and imputing missing values, and (iv) label-encoding remaining categoricals. 
We then assign each sample a scalar difficulty $L_i$ by training lightweight per-target predictors on $V$ and aggregating their per-sample errors: absolute error for regression and soft error $1-\Pr(\text{correct} \mid x_i)$ for classification. 
Sampling probabilities are defined by a linearly scaled and clipped map $p_i=\mathrm{clip}(\lambda L_i,\, p_{\min},\, 1)$ with a small auto-tuning loop on $\lambda$ to match an expected sampling rate $\tau$ while capping the fraction of saturated points at probability $1$. 
Finally, we draw $z_i \sim \mathrm{Bernoulli}(p_i)$ independently and return the full-fidelity subset $D_{\text{sub}}=D[\{i:z_i=1\}]$.
This emphasizes informative (higher-loss) regions while preserving exploration via the floor $p_{\min}$.

The full results are shown in Tables~\ref{tab:results_regressor_xgb} and \ref{tab:results_lgbm_rf} (no sampling);
\ref{tab:results_frac0.5_regressor_xgb} and \ref{tab:results_frac0.5_lgbm_rf} (fraction = 0.5);
\ref{tab:results_frac0.75_regressor_xgb} and \ref{tab:results_frac0.75_lgbm_rf} (fraction = 0.75);
and \ref{tab:results_frac1_regressor_xgb} and \ref{tab:results_frac1_lgbm_rf} (fraction = 1.0).
The timing overhead with and without sample selection is shown in Table \ref{tab:timing_overhead}.

\begin{table*}[ht]
\centering
\caption{Hypervolume and Spread Results (No Sampling): TabNet Regressor and TabNet Embedding + XGBoost}
\label{tab:results_regressor_xgb}
\resizebox{\textwidth}{!}{%
\begin{tabular}{l|cccc|cccc}
\hline
\multirow{2}{*}{Dataset / Metric} & \multicolumn{4}{c|}{TabNet Regressor} & \multicolumn{4}{c}{TabNet Embedding + XGBoost} \\
 & MOBO & SOBO (Runtime) & SOBO (Power) & Random & MOBO & SOBO (Runtime) & SOBO (Power) & Random \\
\hline
PM100 -- Hypervolume & 8.58E+03 & 5.50E+03 & 9.41E+03 & 2.87E+09 & 0.00E+00 & 2.56E+08 & 2.56E+08 & 4.65E+08 \\
PM100 -- Spread      & 9.60E+04 & 8.40E+03 & 2.57E+05 & 9.53E+04 & 0.00E+00 & 1.34E+05 & 1.34E+05 & 3.21E+05 \\
Adastra -- Hypervolume & 3.92E+06 & 3.91E+06 & 3.91E+06 & 1.06E+14 & 0.00E+00 & 4.81E+12 & 4.81E+12 & 4.97E+11 \\
Adastra -- Spread      & 1.42E+05 & 2.61E+04 & 1.43E+05 & 1.42E+05 & 0.00E+00 & 2.57E+07 & 2.57E+07 & 3.79E+07 \\
\hline
\end{tabular}%
}
\end{table*}

\begin{table*}[ht]
\centering
\caption{Hypervolume and Spread Results (No Sampling): TabNet Embedding + LightGBM and RF}
\label{tab:results_lgbm_rf}
\resizebox{\textwidth}{!}{%
\begin{tabular}{l|cccc|cccc}
\hline
\multirow{2}{*}{Dataset / Metric} & \multicolumn{4}{c|}{TabNet Embedding + LightGBM} & \multicolumn{4}{c}{TabNet Embedding + RF} \\
 & MOBO & SOBO (Runtime) & SOBO (Power) & Random & MOBO & SOBO (Runtime) & SOBO (Power) & Random \\
\hline
PM100 -- Hypervolume & 0.00E+00 & 2.73E+08 & 2.73E+08 & 4.51E+08 & 0.00E+00 & 2.61E+08 & 2.61E+08 & 4.52E+08 \\
PM100 -- Spread      & 0.00E+00 & 1.43E+05 & 1.43E+05 & 3.29E+05 & 0.00E+00 & 1.35E+05 & 1.35E+05 & 3.24E+05 \\
Adastra -- Hypervolume & 0.00E+00 & 6.50E+12 & 6.50E+12 & 3.94E+12 & 1.16E+04 & 5.96E+12 & 5.96E+12 & 6.09E+12 \\
Adastra -- Spread      & 0.00E+00 & 3.48E+07 & 3.48E+07 & 3.54E+07 & 7.98E+04 & 3.15E+07 & 3.15E+07 & 3.71E+07 \\
\hline
\end{tabular}%
}
\end{table*}

\begin{table*}[ht]
\centering
\caption{Hypervolume and Spread Results with Intelligent Sample Selection (fraction = 0.5): TabNet Regressor and TabNet Embedding + XGBoost}
\label{tab:results_frac0.5_regressor_xgb}
\resizebox{\textwidth}{!}{%
\begin{tabular}{l|cccc|cccc}
\hline
\multirow{2}{*}{Dataset / Metric} & \multicolumn{4}{c|}{TabNet Regressor} & \multicolumn{4}{c}{TabNet Embedding + XGBoost} \\
 & MOBO & SOBO (Runtime) & SOBO (Power) & Random & MOBO & SOBO (Runtime) & SOBO (Power) & Random \\
\hline
PM100 -- Hypervolume   & 1.34E+10 & 1.34E+10 & 1.30E+10 & 2.05E+12 & 0.00E+00 & 2.73E+08 & 2.73E+08 & 4.65E+11 \\
PM100 -- Spread        & 1.60E+08 & 3.69E+06 & 3.78E+06 & 1.28E+08 & 0.00E+00 & 1.41E+05 & 1.41E+05 & 4.65E+11 \\
Adastra -- Hypervolume & 1.60E+06 & 1.43E+06 & 6.39E+04 & 6.13E+13 & 0.00E+00 & 6.49E+12 & 6.49E+12 & 4.20E+12 \\
Adastra -- Spread      & 2.79E+04 & 5.76E+03 & 5.67E+03 & 2.85E+04 & 0.00E+00 & 3.48E+07 & 3.48E+07 & 2.00E+07 \\
\hline
\end{tabular}%
}
\end{table*}

\begin{table*}[ht]
\centering
\caption{Hypervolume and Spread Results with Intelligent Sample Selection (fraction = 0.5): TabNet Embedding + LightGBM and RF}
\label{tab:results_frac0.5_lgbm_rf}
\resizebox{\textwidth}{!}{%
\begin{tabular}{l|cccc|cccc}
\hline
\multirow{2}{*}{Dataset / Metric} & \multicolumn{4}{c|}{TabNet Embedding + LightGBM} & \multicolumn{4}{c}{TabNet Embedding + RF} \\
 & MOBO & SOBO (Runtime) & SOBO (Power) & Random & MOBO & SOBO (Runtime) & SOBO (Power) & Random \\
\hline
PM100 -- Hypervolume   & 0.00E+00 & 2.51E+08 & 2.51E+08 & 4.61E+11 & 0.00E+00 & 2.75E+08 & 2.75E+08 & 4.56E+11 \\
PM100 -- Spread        & 0.00E+00 & 1.38E+05 & 1.38E+05 & 1.02E+06 & 0.00E+00 & 1.42E+05 & 1.42E+05 & 9.99E+05 \\
Adastra -- Hypervolume & 0.00E+00 & 6.18E+12 & 6.18E+12 & 4.69E+12 & 1.43E+07 & 6.26E+12 & 6.26E+12 & 4.21E+12 \\
Adastra -- Spread      & 0.00E+00 & 3.32E+07 & 3.32E+07 & 2.08E+07 & 1.44E+04 & 3.35E+07 & 3.35E+07 & 1.99E+07 \\
\hline
\end{tabular}%
}
\end{table*}

\begin{table*}[ht]
\centering
\caption{Hypervolume and Spread Results with Intelligent Sample Selection (fraction = 0.75): TabNet Regressor and TabNet Embedding + XGBoost}
\label{tab:results_frac0.75_regressor_xgb}
\resizebox{\textwidth}{!}{%
\begin{tabular}{l|cccc|cccc}
\hline
\multirow{2}{*}{Dataset / Metric} & \multicolumn{4}{c|}{TabNet Regressor} & \multicolumn{4}{c}{TabNet Embedding + XGBoost} \\
 & MOBO & SOBO (Runtime) & SOBO (Power) & Random & MOBO & SOBO (Runtime) & SOBO (Power) & Random \\
\hline
PM100 -- Hypervolume   & 1.15E+03 & 1.15E+03 & 1.15E+03 & 2.58E+09 & 0.00E+00 & 2.49E+08 & 2.49E+08 & 4.69E+11 \\
PM100 -- Spread        & 3.34E+02 & 7.73E+01 & 2.86E+02 & 3.31E+02 & 0.00E+00 & 1.29E+05 & 1.29E+05 & 9.87E+05 \\
Adastra -- Hypervolume & 1.05E+02 & 1.05E+02 & 1.05E+02 & 1.54E+16 & 0.00E+00 & 6.02E+12 & 6.02E+12 & 4.17E+13 \\
Adastra -- Spread      & 4.82E+02 & 4.82E+02 & 4.82E+02 & 0.00E+00 & 0.00E+00 & 3.22E+07 & 3.22E+07 & 4.75E+07 \\
\hline
\end{tabular}%
}
\end{table*}

\begin{table*}[ht]
\centering
\caption{Hypervolume and Spread Results with Intelligent Sample Selection (fraction = 0.75): TabNet Embedding + LightGBM and RF}
\label{tab:results_frac0.75_lgbm_rf}
\resizebox{\textwidth}{!}{%
\begin{tabular}{l|cccc|cccc}
\hline
\multirow{2}{*}{Dataset / Metric} & \multicolumn{4}{c|}{TabNet Embedding + LightGBM} & \multicolumn{4}{c}{TabNet Embedding + RF} \\
 & MOBO & SOBO (Runtime) & SOBO (Power) & Random & MOBO & SOBO (Runtime) & SOBO (Power) & Random \\
\hline
PM100 -- Hypervolume   & 0.00E+00 & 2.70E+08 & 2.70E+08 & 4.58E+11 & 0.00E+00 & 2.78E+08 & 2.78E+08 & 4.56E+11 \\
PM100 -- Spread        & 0.00E+00 & 1.36E+05 & 1.36E+05 & 1.01E+06 & 0.00E+00 & 1.44E+05 & 1.44E+05 & 9.99E+05 \\
Adastra -- Hypervolume & 0.00E+00 & 6.34E+12 & 6.34E+12 & 2.55E+13 & 3.44E+03 & 6.26E+12 & 6.26E+12 & 3.56E+13 \\
Adastra -- Spread      & 0.00E+00 & 3.40E+07 & 3.40E+07 & 4.03E+07 & 3.39E+04 & 4.77E+07 & 4.77E+07 & 3.33E+07 \\
\hline
\end{tabular}%
}
\end{table*}

\begin{table*}[ht]
\centering
\caption{Hypervolume and Spread Results with Intelligent Sample Selection (fraction = 1): TabNet Regressor and TabNet Embedding + XGBoost}
\label{tab:results_frac1_regressor_xgb}
\resizebox{\textwidth}{!}{%
\begin{tabular}{l|cccc|cccc}
\hline
\multirow{2}{*}{Dataset / Metric} & \multicolumn{4}{c|}{TabNet Regressor} & \multicolumn{4}{c}{TabNet Embedding + XGBoost} \\
 & MOBO & SOBO (Runtime) & SOBO (Power) & Random & MOBO & SOBO (Runtime) & SOBO (Power) & Random \\
\hline
PM100 -- Hypervolume   & 4.29E+04 & 3.47E+04 & 4.36E+04 & 9.25E+10 & 1.38E+02 & 2.55E+08 & 2.56E+08 & 4.63E+11 \\
PM100 -- Spread        & 4.10E+05 & 2.17E+04 & 4.72E+05 & 3.91E+05 & 2.65E+03 & 1.30E+05 & 1.30E+05 & 8.04E+05 \\
Adastra -- Hypervolume & 4.01E-02 & 2.93E-02 & 2.93E-02 & 2.49E+13 & 3.31E+03 & 1.18E+13 & 1.18E+13 & 1.24E+14 \\
Adastra -- Spread      & 4.70E-01 & 4.70E-01 & 4.70E-01 & 0.00E+00 & 4.03E+04 & 6.31E+07 & 6.31E+07 & 5.62E+07 \\
\hline
\end{tabular}%
}
\end{table*}

\begin{table*}[ht]
\centering
\caption{Hypervolume and Spread Results with Intelligent Sample Selection (fraction = 1): TabNet Embedding + LightGBM and RF}
\label{tab:results_frac1_lgbm_rf}
\resizebox{\textwidth}{!}{%
\begin{tabular}{l|cccc|cccc}
\hline
\multirow{2}{*}{Dataset / Metric} & \multicolumn{4}{c|}{TabNet Embedding + LightGBM} & \multicolumn{4}{c}{TabNet Embedding + RF} \\
 & MOBO & SOBO (Runtime) & SOBO (Power) & Random & MOBO & SOBO (Runtime) & SOBO (Power) & Random \\
\hline
PM100 -- Hypervolume   & 1.38E+02 & 2.84E+08 & 2.84E+08 & 4.30E+11 & 0.00E+00 & 2.70E+08 & 2.70E+08 & 4.55E+11 \\
PM100 -- Spread        & 0.00E+00 & 1.46E+05 & 1.46E+05 & 7.81E+05 & 0.00E+00 & 1.40E+05 & 1.40E+05 & 8.12E+05 \\
Adastra -- Hypervolume & 3.31E+03 & 1.99E+13 & 1.99E+13 & 1.91E+14 & 1.58E+04 & 6.54E+12 & 6.54E+12 & 8.23E+12 \\
Adastra -- Spread      & 3.30E+00 & 1.08E+08 & 1.08E+08 & 9.10E+07 & 6.42E+05 & 3.50E+07 & 3.50E+07 & 3.69E+07 \\
\hline
\end{tabular}%
}
\end{table*}

\begin{table*}[ht]
\centering
\caption{Comparison of timing overhead (seconds) for PM100 and Adastra datasets with and without intelligent sampling (fraction = 1).}
\label{tab:timing_overhead}
\resizebox{\textwidth}{!}{%
\begin{tabular}{l|cc|cc}
\hline
\multirow{2}{*}{Step} & \multicolumn{2}{c|}{PM100} & \multicolumn{2}{c}{Adastra} \\
 & No Sampling & With Sampling (Fraction = 1) & No Sampling & With Sampling (Fraction = 1) \\
\hline
Preprocessing      & 10.82   & 6.36   & 0.80    & 0.23   \\
Runtime Model      & 2359.40 & 1264.97 & 237.73  & 68.02  \\
Power Model        & 2002.76 & 959.08  & 183.99  & 53.62  \\
Preproc. MOBO      & 10.20   & 6.53    & 0.64    & 0.23   \\
MOBO               & 1883.77 & 3633.60 & 2065.70 & 1642.01 \\
SOBO Runtime       & 104.86  & 71.11   & 120.39  & 95.58  \\
SOBO Power         & 108.97  & 67.81   & 90.18   & 101.99 \\
\hline
TOTAL              & 6480.78 & 6009.46 & 2699.43 & 1961.68 \\
\hline
\end{tabular}%
}
\end{table*}

\begin{table*}[t]
\centering
\caption{Common experimental settings used across both pipelines (Adastra; PM100 uses the same settings unless noted).}
\label{tab:common_settings}
\begin{tabularx}{\textwidth}{l X}
\toprule
\textbf{Setting} & \textbf{Value / Notes} \\
\midrule
Datasets & Adastra (15 days log trace); PM100 (SC-W'23) \\
Sampling fractions & 0.5,\; 0.75,\; 1.0 (active-learning guided) \\
Design variable & \texttt{num\_nodes\_alloc} \\
MOBO acquisition & \texttt{qLogExpectedHypervolumeImprovement} (\texttt{logEHVI}) \\
MOBO iterations & 300 (q=1 candidate/step; \texttt{optimize\_acqf}: 5 restarts, 32 raw samples) \\
GP model & Multi-output GP (BoTorch), reference point inferred online \\
Baselines & SOBO (runtime-only), SOBO (power-only), Random (5 seeds; split budget) \\
Random baseline & \texttt{N\_RANDOM\_SEEDS}=5; total points $\approx$ MOBO budget (evenly split per seed) \\
Figure naming & “Transformer-based model” in figures denotes TabNet regressor (text aligned accordingly) \\
\bottomrule
\end{tabularx}
\end{table*}

\begin{table*}[t]
\centering
\caption{Pipeline-specific training and model hyperparameters. PM100 uses the same settings; only the dataset changes.}
\label{tab:pipeline_hparams}
\begin{tabularx}{\textwidth}{l X X}
\toprule
\textbf{Component} & \textbf{Regressor Pipeline (TabNet only)} & \textbf{Embeddings Pipeline (TabNet + RF/XGB/LGBM)} \\
\midrule
\multicolumn{3}{l}{\emph{TabNet training (fit parameters)}} \\
\midrule
Max epochs & 1000 & 100 \\
Early stopping & Patience = 20 & Patience = 10 \\
Batch size & 1024 & 1024 \\
Virtual batch size & 128 & 128 \\
Eval metric & MSE (valid) & RMSE (valid) \\
Seed & \texttt{seed} (set in notebook) & \texttt{random\_state}=42 (downstream); TabNet uses default seed unless overridden \\
Target transforms & Robust/log transforms supported; inverse-transform for eval & Same; targets transformed once, meta stored for inverse transforms \\
\midrule
\multicolumn{3}{l}{\emph{Downstream regressors (used only in embeddings pipeline)}} \\
\midrule
Random Forest & --- & \texttt{n\_estimators}=100, \texttt{max\_depth}=10, \texttt{random\_state}=42, \texttt{n\_jobs}=-1 \\
XGBoost & --- & \texttt{n\_estimators}=100, \texttt{max\_depth}=6, \texttt{learning\_rate}=0.10, \texttt{random\_state}=42 \\
LightGBM & --- & \texttt{n\_estimators}=100, \texttt{max\_depth}=6, \texttt{learning\_rate}=0.10, \texttt{random\_state}=42 \\
\midrule
\multicolumn{3}{l}{\emph{MOBO loop (both pipelines)}} \\
\midrule
Acquisition & \multicolumn{2}{X}{\texttt{qLogExpectedHypervolumeImprovement} (logEHVI), Sobol sampler with 128 normal samples} \\
Optimization & \multicolumn{2}{X}{\texttt{optimize\_acqf}: 5 restarts, 32 raw samples; $q=1$} \\
Reference point & \multicolumn{2}{X}{Inferred dynamically from current $Y$ (minimization space with negated objectives)} \\
Pareto metrics & \multicolumn{2}{X}{Hypervolume, Spread (raw minimization space; signs handled consistently)} \\
\bottomrule
\end{tabularx}
\end{table*}

\begin{table}[h]
\centering
\small
\caption{Effect of Active Learning (sampling fractions) on surrogate accuracy and optimization metrics.}
\label{tab:effect}
\begin{tabularx}{\linewidth}{l c c c X}
\toprule
\textbf{Dataset} & \textbf{Fraction} & \textbf{Train Size} & \textbf{MAPE} & \textbf{Observed Effect (HV/Spread)} \\
\midrule
\multirow{3}{*}{PM100} 
  & 50\%  & 73,983  & $\approx$0.99 & HV stable; Spread drops from $2.5\times10^5$ to $<10^4$ \\
  & 75\%  & 77,278  & $\approx$0.99 & Accuracy plateaus; HV/Spread stable \\
  & 100\% & 109,202 & $\approx$0.99 & No further gains vs.\ 75\% \\
\midrule
\multirow{3}{*}{Adastra} 
  & 50\%  & 3,964   & $\approx$0.99 & HV stable; erratic values ($10^{16}\!\to\!10^{13}$) suppressed \\
  & 75\%  & 4,163   & $\approx$0.99 & Accuracy plateaus; smoother trade-offs \\
  & 100\% & 4,547   & $\approx$0.99 & No further gains vs.\ 75\% \\
\bottomrule
\end{tabularx}
\end{table}

\begin{figure}[t]
  \centering
  \includegraphics[width=\linewidth]{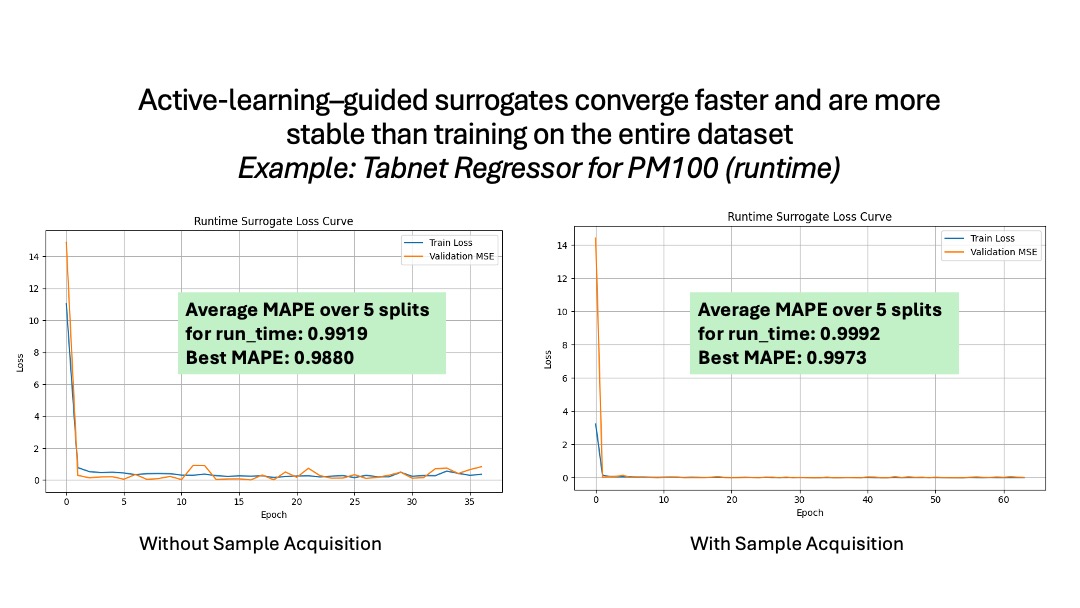}
  \caption{Runtime surrogate convergence across different models.}
  \label{fig:active2}
\end{figure}

\begin{figure}[t]
  \centering
  \includegraphics[width=\linewidth]{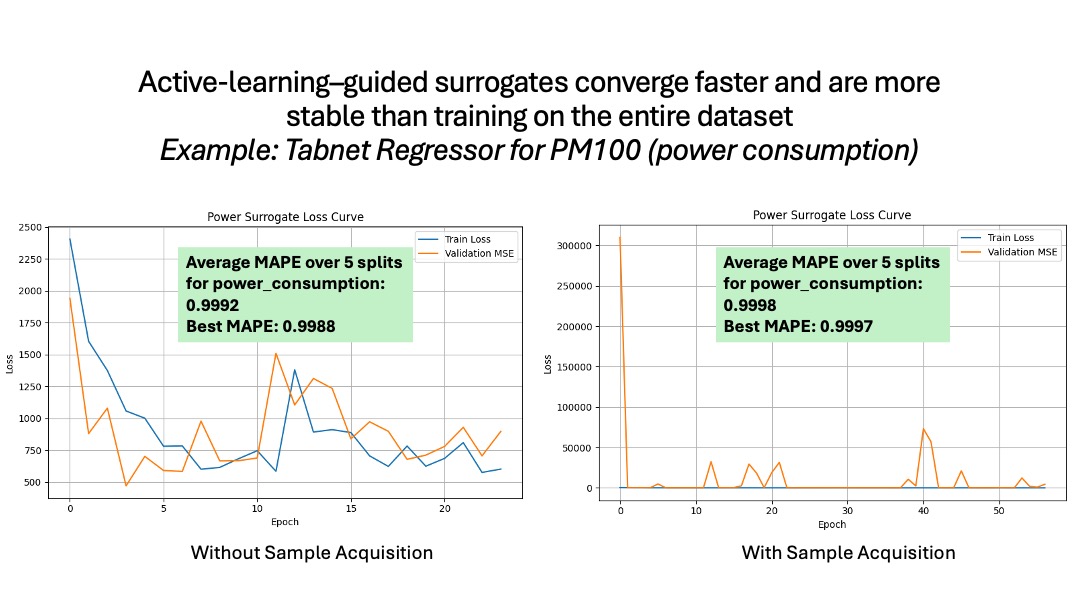}
  \caption{Power surrogate convergence across different models.}
  \label{fig:active}
\end{figure}

The figures \ref{fig:active2} and \ref{fig:active} show the convergence of surrogates across different models.

\begin{figure}[t]
  \centering
  \includegraphics[width=\linewidth]{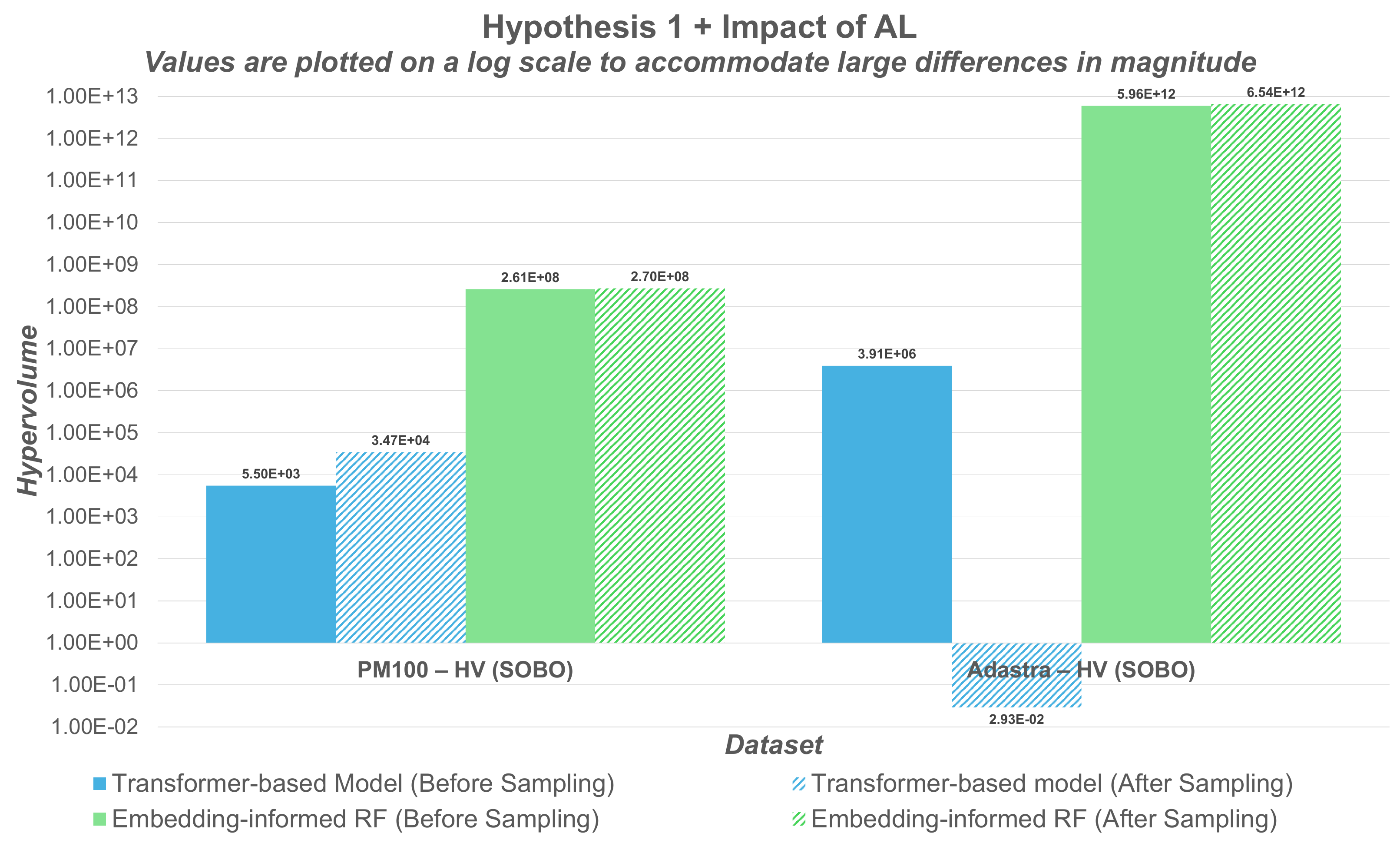}
  \caption{Embedding-informed surrogates yield higher HV than TabNet models on PM100 and Adastra.}
  \label{fig:h1}
\end{figure}

\begin{figure}[t]
  \centering
  \includegraphics[width=\linewidth]{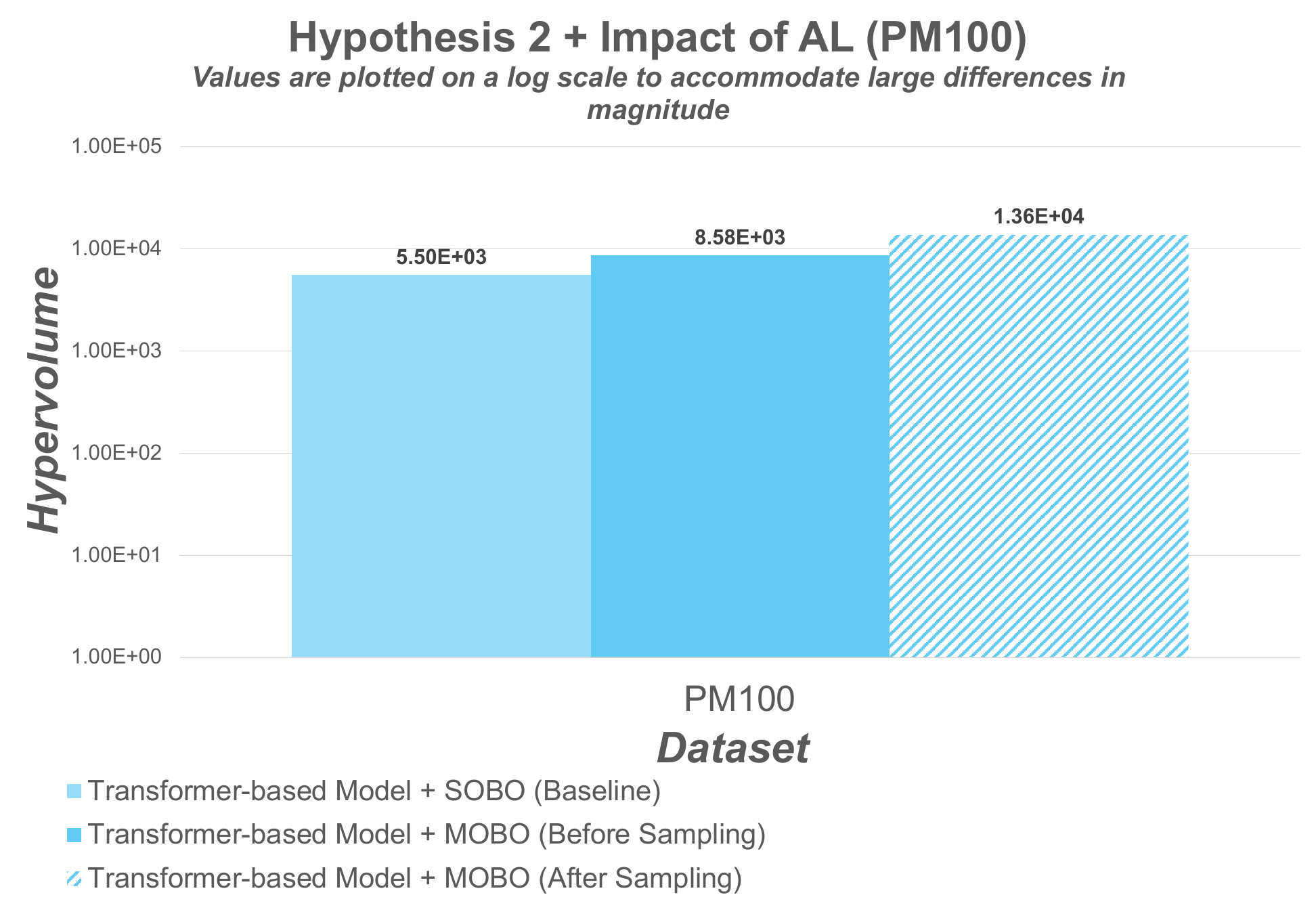}
  \caption{MOBO achieves higher HV than SOBO on PM100, revealing better runtime–power trade-offs.}
  \label{fig:h2_p1}
\end{figure}

\begin{figure}[t]
  \centering
  \includegraphics[width=\linewidth]{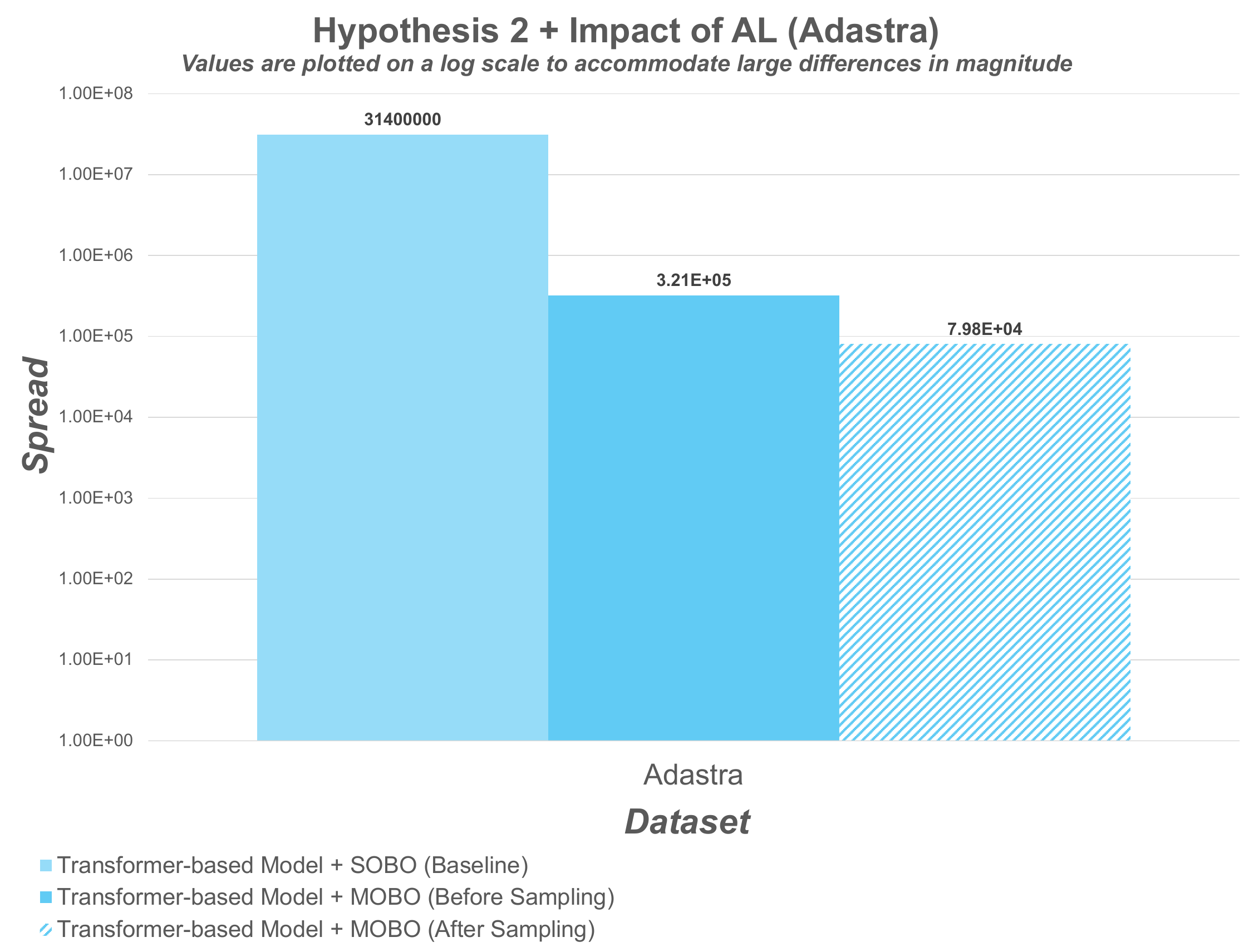}
  \caption{MOBO achieves lower Spread than SOBO on Adastra, indicating more balanced Pareto fronts.}
  \label{fig:h2_p2}
\end{figure}

The figures \ref{fig:h1}, \ref{fig:h2_p1}, and \ref{fig:h2_p2} show evaluation results across two HPC datasets, showing surrogate quality and trade-off performance,

\paragraph{Licensing of Assets.}  
The PM100 dataset [10] is distributed under the \textbf{Creative Commons Attribution 4.0 (CC BY 4.0)} license, as noted in the SC’23 proceedings.  
The Adastra system documentation and log data [11] are publicly described by CINES/GENCI, but no explicit license is specified; we assume standard institutional terms for research use.  
The Stampede3 user guide [12] is published on the TACC documentation portal and made available under TACC’s usage policy; no formal open license is displayed.  

\clearpage


\end{document}